\documentclass{sigchi}

\usepackage{etoolbox}
\makeatletter
\patchcmd{\maketitle} {\@copyrightspace}{}{}{}
\makeatother

\usepackage{subcaption}
\usepackage{graphicx}
\usepackage{caption, float}
\usepackage{subcaption}
\newcommand{\rulesep}{\unskip\ \vrule\ }
\usepackage{lipsum}

\newcommand\blfootnote[1]{%
  \begingroup
  \renewcommand\thefootnote{}\footnote{#1}%
  \addtocounter{footnote}{-1}%
  \endgroup
}

\usepackage{balance}       % to better equalize the last page
\usepackage{graphics}      % for EPS, load graphicx instead 
\usepackage[T1]{fontenc}   % for umlauts and other diaeresis
\usepackage{txfonts}
\usepackage{mathptmx}
\usepackage[pdflang={en-US},pdftex]{hyperref}
\usepackage{color}
\usepackage{booktabs}
\usepackage{textcomp}

\usepackage{microtype}        % Improved Tracking and Kerning
\usepackage{ccicons}          % Cite your images correctly!
\usepackage{todonotes}

\def\plaintitle{Can You be More Social? Injecting Politeness and Positivity into Task-Oriented Conversational Agents}
\def\plainauthor{First Author, Second Author, Third Author,
  Fourth Author, Fifth Author, Sixth Author}

\def\plainkeywords{Authors' choice; of terms; separated; by
  semicolons; include commas, within terms only; this section is required.}

\makeatletter
\def\url@leostyle{%
  \@ifundefined{selectfont}{
    \def\UrlFont{\sf}
  }{
    \def\UrlFont{\small\bf\ttfamily}
  }}
\makeatother
\def\pprw{8.5in}
\def\pprh{11in}

\definecolor{linkColor}{RGB}{6,125,233}
\hypersetup{%
  pdftitle={\plaintitle},
  pdfauthor={\plainauthor},
  pdfkeywords={\plainkeywords},
  pdfdisplaydoctitle=true, % For Accessibility
  bookmarksnumbered,
  pdfstartview={FitH},
  colorlinks,
  citecolor=black,
  filecolor=black,
  linkcolor=black,
  urlcolor=linkColor,
  breaklinks=true,
  hypertexnames=false
}

\begin{document}

\title{\plaintitle}

\numberofauthors{6}
\author{%
  \alignauthor{Yi-Chia Wang\\
    \affaddr{Uber AI}\\
    \email{yichia.wang@gmail.com}}\\
  \alignauthor{Alexandros Papangelis\\
    \affaddr{Uber AI}\\
    \email{al3x.papangelis@gmail.com}}\\
  \alignauthor{Runze Wang\\
    \affaddr{Uber AI}\\
    \email{ernuzwang@gmail.com}}\\
  \alignauthor{Zhaleh Feizollahi\\
    \affaddr{Uber AI}\\
    \email{zhaleh.feizollahi@gmail.com}}\\
  \alignauthor{Gokhan Tur\\
    \affaddr{Uber AI}\\
    \email{gokhan.tur@ieee.org}}\\
  \alignauthor{Robert Kraut\\
    \affaddr{Human-Computer Interaction Institute}\\
    \affaddr{Carnegie Mellon University}\\
    \email{robert.kraut@cmu.edu}}\\
}

\maketitle

\begin{abstract}
Goal-oriented conversational agents are becoming prevalent in our daily lives. For these systems to engage users and achieve their goals, they need to exhibit appropriate social behavior as well as provide informative replies that guide users through tasks.  The first component of the research in this paper applies statistical modeling techniques to understand conversations between users and human agents for customer service.  Analyses show that social language used by human agents is associated with greater users' responsiveness and task completion. The second component of the research is the construction of a conversational agent model capable of injecting social language into an agent's responses while still preserving content.  The model uses a sequence-to-sequence deep learning architecture, extended with a social language understanding element. Evaluation in terms of content preservation and social language level using both human judgment and automatic linguistic measures shows that the model can generate responses that enable agents to address users' issues in a more socially appropriate way. 
\blfootnote{This work was done at Uber AI.  It is a revision of our work presented in the Second Workshop on Conversational AI @ NeurIPS 2018 titled "Can You be More Polite and Positive? Infusing Social Language into Task-Oriented Conversational Agents" which is non-archival.}
\end{abstract}

% ACM Classfication



% Author Keywords

% Print the classficiation codes
%Please use the 2012 Classifiers and see this link to embed them in the text: \url{https://dl.acm.org/ccs/ccs_flat.cfm}

\section{Background and Introduction}

Conversational agents are becoming part of our lives.  These systems generally fall into two categories, \textit{task-oriented assistants} and \textit{chatbots} \cite{Shum2018}.  Task-oriented assistants are designed to fulfill a specific task by having single-turn or multi-turn conversations with users to retrieve information from them (e.g. \textit{Microsoft Cortana}, \textit{Apple Siri}, \textit{Google Assistant}).  Chatbots are designed to have more socially-oriented chit chat with users. The goal is usually to mimic human-human conversations and engage users in those conversations for as long as possible (e.g. \textit{ELIZA}\cite{Weizenbaum:1966:ECP:365153.365168},  \textit{XiaoIce}\cite{zhou2018design}, or \textit{TickTock}\cite{yu2015ticktock}).  

In order to have extended human-like conversations, some researchers have studied how to incorporate social language into chatbots to generate proper interpersonal responses and build an emotional connection with users \cite{Shum2018}. For example, \textit{XiaoIce}, Microsoft's social chatbot in China, can respond with empathetic language and show care while chatting with users.  However, there are only a handful of studies that focus on incorporating social capabilities into task-oriented assistants \cite{Bickmore:2001:RAM:365024.365304, Cassell2003, Bickmore:2009:TTC:1518701.1518891, walker1997improvising, brixey2019building} even though prior literature has suggested that these factors might play an important role in the process of task-oriented conversations and be associated with better user engagement and satisfaction \cite{1545303, Liao:2016:YSS:2901790.2901842, Bickmore:2005:EML:1067860.1067867, Bickmore:2009:TTC:1518701.1518891}.  Thus, in this work we try to answer the following research questions: (1) Does social language used by humans in task-oriented conversations affect user responsiveness and task completion? If so, how? (2) Can we effectively introduce social aspects of language in the responses of a task-oriented conversational agent?

We focus on the customer service domain in which customer service personnel help drivers sign up with a ride-sharing provider, since customer service is a typical application area for automated task-oriented assistants  Moreover, driver on-boarding support is a closed-domain problem and a well-defined task.  We first describe an empirical study to quantitatively examine the relationship of customer service representatives' use of social language  to drivers' responsiveness and the completion of their first trip, based on an analysis of driver and human-agent conversations.  The social language is measured using existing pre-trained machine learning models developed in prior literature.  After that, we apply the findings to build an end-to-end deep learning model to generate automated agent responses given driver inquiries.  Our aim is to train a task-oriented agent that can produce dialogues with the desired level of social language while still maintaining the necessary content to guide drivers through the on-boarding process and lead them to complete their first trip.  The main contributions of this work are below:

\begin{enumerate}
\item Systematically analyze the relationship between social language and user responsiveness as well as task completion in a large real-world conversational dataset.
\item Propose a deep learning framework for task-oriented dialogue generation that includes a social language understanding and production component.
\item Use human judgment and automatic assessment methods to evaluate the extent to which the new dialogue generation model preserves task-oriented content and incorporates social language generation.
\end{enumerate}

\section{Related Work}

In 1978, Bloom and Lahey proposed a language development framework that suggests that language has three components: \textit{content}, \textit{form}, and \textit{use} \cite{bloom1978language}.  \textit{Content} refers to semantics and underlying meaning in text; \textit{form} is related to the rules of language such as morphology and syntax.  The \textit{use} of language,  also called \textit{pragmatics} or \textit{social language}, refers to how language is used and interpreted in different social settings or contexts.  They point out that social language is important for various interpersonal functions.  Humans use a variety of social language strategies to maintain and develop interpersonal relationships, such as increasing intimacy through self-disclosure \cite{collins1994self} and building common ground through small talk \cite{clark1996}.  The current research focuses on better understanding the use of social language in a human-to-human, task-oriented context and proposes a architecture to computationally generate such language. We start by reviewing the related work on social language generation in task-oriented assistants and then summarize the literature on politeness and positivity that motivates our work.

\subsection{Social Language in Task-Oriented Conversational Agents}

Despite the importance of social language for human-human relationships and interactions, most task-oriented conversational assistants only focus on presenting the right content to users; nevertheless, a few have tried to incorporate social language, with most work in this direction done in the context of embodied conversational agents.  Embodied agents have a physical or graphical representation of a body or face and are usually capable of interacting with users through multiple modalities (e.g. linguistic content, tone of voice, gestures, facial expression). Human social communication and interaction is a complex process which often involves coordination across the different modalities. Some research has examined how to effectively facilitate this coordination.  For example, Bickmore and Cassell integrated a theory of social dialogue in a real estate conversational agent (REA) and demonstrated that small talk can help the virtual agent build trust with users \cite{Cassell2003, Bickmore:2001:RAM:365024.365304}. However, the REA was not fully automated but controlled by a human wizard who followed scripts during the experiment. Following up on this work, Zhao et al. \cite{zhao2014towards} proposed a theoretical framework and a computational model \cite{papangelis2014towards} of how humans use various conversational strategies to build, maintain, or break rapport. This framework was used to develop SARA \cite{matsuyama2016socially}, an embodied conversational agent able to build rapport with humans and perform tasks such as calendar scheduling. In addition to REA and SARA, examples of embodied agents that use social language to achieve a task include Greta \cite{bevacqua2007expressive, niewiadomski2008expressions}, Ellie \cite{devault2014simsensei}, Zara \cite{siddique2017zara}, Nora \cite{winata2017nora} and other virtual \cite{gratch2007creating, paranjape2018towards, d2012autotutor} or robotic agents \cite{leite2013social}.  However, because many interactions with virtual agents now occur online via text, it is important to understand how to encode and generate social language, especially when text is the only available modality. 

In many commercial systems, templates, rules, or content filters are used to generate social norms of language. The recent Alexa Prize competitions \cite{ram2018conversational, khatri2018alexa} have fostered interesting work that requires an assistant to fuse task-oriented goals (e.g. play music) with social talk. However, because of various challenges related to deploying agents to real customers, none of these efforts use statistical methods to explicitly model social aspects of language generation.  Instead, most attempts are template-based or script-based.

When considering approaches for generating social language in task-oriented conversational agents, most attempts are template-based or script-based, which interleave task-related utterances with socially-related utterances.  Zhao et al. \cite{zhao2018sogo} proposed to alternate task-related utterances (``Task Phase'' - negotiation) with utterances related to social strategies (``Social Phase''), for example expressing gratitude or self-disclosure.  Chandar et al. \cite{chandar2017leveraging} proposed an agent to assist with an onboarding task (e.g. by providing information, searching, or proactively reminding potential employees of pending tasks) and including chit-chat capabilities. However, this agent did not model social language directly, but used a neural network to select the most appropriate response from a pre-defined list. Our work differs from these by building an end-to-end deep neural network to jointly understand input utterances and generate output utterances infused with social language, instead of having separate modules for recognition and understanding of social features and for language generation.

\subsection{Positivity and Politeness in Task Completion}

Gnewuch and colleagues \cite{gnewuch2017towards} proposed twelve design principles for social conversational agents, according to which conversational agents should not only adhere to Grice's maxims \cite{grice1975logic} but also be responsive to social cues since humans expect an experience that resembles human communication \cite{moon2000intimate, moon2003don, nass2000machines, knijnenburg2016inferring}. In particular, they state that conversational agents \emph{``[...] should produce social cues (e.g. appearance or language) that correspond to these service agent characteristics as well as fit to the context in which they are implemented.''} \cite[p. 8]{gnewuch2017towards}. Because our agent is text-based, we focus on natural language only. Although there are  many variants of social language, we  concentrate here on \textit{politeness} and \textit{positivity} because prior research suggests that these two types of social language strategies can be important for more natural human-machine conversations and customer service interactions, as described below.  

According to politeness theory \cite{levinson1987politeness}, politeness is a common social language strategy used for saving ``face''.  It helps regulate the social distance between two parties and removes face threats, which may lead to feeling awkward or embarrassed. Thus, the ability of a conversational agent to respond in a polite manner can protect users from ``losing face''. Politeness has also been shown to lead to the development of trust \cite{svennevig2000getting,Bickmore:2001:RAM:365024.365304} and rapport \cite{spencer2005politeness, tickle1990nature}, which in turn leads to better communication and performance in collaborative tasks \cite{bernieri1991interpersonal,drolet2000rapport,kang2012towards,garbarski2016interviewing,nadler2003rapport}. 

Positivity is "the quality or state of being positive" \cite{positivity_nd}. Researchers have shown that positivity can engage employees and improve their performance in the work place \cite{sweetman2010power}.  Positivity is also contagious \cite{kramer2014experimental} and leads to likability \cite{dainton1994maintenance}. So we argue that when users work with a conversational agent to accomplish a task, they would also like to talk to one which exhibits more positive behavior or uses more positive language.

We, therefore, hypothesize that users would prefer a conversational agent with more polite or more positive language and be more willing to engage with, respond to and persist in the interaction when conversing with an agent using polite and positive language. In the context of a task in which ride-share drivers are registering, this type of language will lead to more drivers completing tasks required for on-boarding and more completing their first trip. Appropriateness of social language and communication strategies such as positivity and politeness, however, may vary according to context and over time \cite{johnstone1989linguistic,giles1991,spencer2005politeness,tickle1990nature}, meaning that simply designing polite language templates may not be enough. Our model addresses this by taking into account the context of the conversation and completed tasks so far and generates language with the desired content and social language norms.

\section{Research Context: Driver On-boarding}

A ride-sharing company provides rides for real-time requests.  Before drivers can start to take passengers, they need to go through an on-boarding process.  Our data originates from an initiative in a large ride-sharing provider where new drivers who created an account to start the on-boarding process are paired with a customer support representative (CSR) who guides them via Short Message Service (SMS) exchanges.  Typically, the driver must complete a series of tasks: consenting to a background check, providing proof of registration and insurance, a vehicle inspection, providing a photo and taking a city knowledge test. Once all the documents have been submitted, verified, and passed, the driver is approved and he or she can complete his/her first trip, which marks the completion of the onboarding task. The job of the CSR is to guide and encourage the driver through the process by answering general questions, providing status updates, reminding drivers of next steps, and resolving any problems during the on-boarding process. The resulting dialogues are SMS exchanges between a driver and the driver's CSR, which may last several days or weeks.

We collected 4 million on-boarding driver-CSR message pairs that were exchanged between drivers and CSRs in the ride-sharing company.  All the messages were de-identified.  The dataset was used to 1) analyze the impact of politeness and positivity in CSR messages on driver responsiveness and task completion and 2) build conversational models to generate agent responses with social language cues.

\section{Study 1: The Relationship between Social Language and User Engagement}

The goal of this study is to empirically estimate how politeness and positivity in CSR responses predict driver engagement by building statistical analytic models.  We measured politeness and positivity using pre-trained machine learning models.  We operationalized driver engagement in terms of their responsiveness and completion of their first trip.

\subsection{Dependent Variables}
We examined two dependent variables relating to driver engagement:

\subsubsection{Driver responsiveness}
If the form of what CSRs say is relevant to drivers, then the likelihood of drivers' response should be affected by CSR message style.  Based on this assumption, we created a binary variable to measure the \textit{short-term driver engagement}.  It was set to 1 if a driver responded to a CSR message within 24 hours or 0 otherwise.

\subsubsection{Completion of first trip}  
In addition to the short-term driver engagement metric, we also considered the \textit{long-term impact} of CSR responses on drivers' finishing the task, i.e., whether CSRs successfully guided drivers through the on-boarding funnel so that drivers completed their first trip.  Specifically, this binary measure was set to 1 if a driver completed his/her first trip within 7 days after a CSR sent him/her a message or 0 otherwise.

\subsection{Independent and Control Variables}
Given a CSR message, we extracted its politeness and positivity levels as independent variables using off-the-shelf classifiers.  We went through several steps to pre-process and clean CSR messages.  Messages were tokenized with the NLTK toolkit \cite{Bird:2009:NLP:1717171}, and personally identifiable information (PII) such as URLs, email, names, dates, and numbers, were replaced with standard tags.  Afterwards, we utilized pre-trained machine learning models to extract the following features:

\subsubsection{Politeness}
We used a state-of-the-art off-the-shelf politeness machine learning model to measure the politeness level of a CSR message \cite{DBLP:conf/acl/Danescu-Niculescu-MizilSJLP13}.  The model was an SVM classifier trained on a corpus with politeness labels and based on domain-independent lexical and syntactic features identified by politeness theory.  In particular, the features were developed to reflect a series of politeness strategies such as gratitude (e.g. \textit{\textbf{I} really \textbf{appreciate} your help.}), apologizing (e.g. \textit{\textbf{Sorry} to disturb you...}), and please (e.g. \textit{Would you \textbf{please}...}).  Experiments conducted by the authors showed that these linguistically informed features generalize well to interactions between CSRs and drivers.   
The classifier outputs a politeness score between 0 and 1 and performs almost as well as human raters across domains.  Below are a few examples of CSR messages with different levels of politeness scores generated by the classifier:

\begin{itemize}
\item (0.27) \textit{Please download the Partner app to confirm your account: <URL>}

\item (0.43) \textit{Hello <Name>, are you still interested in partnering with us? You're so close to hitting the road and making some money while driving.}

\item (0.97) \textit{Hello, my name is <Name> your Account Specialist. Good news! It looks like your background check has passed!  The final step to earning with us is uploading your registration. Could you please text me a clear photo of your registration so I can upload it to your account?}
\end{itemize}

\subsubsection{Positivity}
Dainton et al. \cite{dainton1994maintenance} defined positivity as language that is upbeat and cheerful. Since there is no existing off-the-shelf model for positivity, we decided to measure positive sentiment as a proxy.  Sentiment refers to the contextual polarity or emotional affect of a text \cite{Wilson:2005:RCP:1220575.1220619}, which is semantically similar to positivity.  To evaluate positive sentiment, we used VADER, a rule-based sentiment analyzer  \cite{DBLP:conf/icwsm/HuttoG14}.  VADER was built with a combination of lexical features and general syntactical and grammatical rules to capture the expression and emphasis of sentiment.  The authors compared its performance with eleven benchmarks including both lexicon-based (e.g. Linguistic Inquiry and Word Count (LIWC) \cite{pennebaker2001linguistic}) and machine learning approaches such as one using a Naive Bayes algorithm.  They showed that VADER outperforms human judges and is generalizable across contexts. Given a piece of text, VADER produces a 3-dimensional measurement to estimate the extent of positive, negative, and neutral sentiment in it.  The three sentiment scores represent the proportion of each sentiment in the text and sum up to one as in the example shown below.  Since positive sentiment score is highly negatively correlated with neutral and negative sentiment, to avoid multicollinearity we only included the positive sentiment score in regression models to predict driver responsiveness and trip completion.

\begin{itemize}
\item (pos=.49, neg=.0, neu=.51) \textit{Nice!  The 2 links I sent you will be your best friends. Good luck! Let me know how it goes for you.}
\end{itemize}

\subsubsection{Control Variables}
In addition to some basic demographic information about drivers such as their age, we also measured the following control variables. By controlling for these, we can make claims about the impact of CSR messages rather than about the driver himself/herself.

\begin{itemize}
\item\textbf{Sign-up city} is a dummy variable controlling for the city where a driver signed up to become a partner with the ride-sharing company.

\item\textbf{Days since signup} is the number of days since the driver registered on the platform.

\item\textbf{Number of previous driver messages} is a measure of how many messages the driver sent to the CSR since he/she signed up.  Different drivers might have different likelihood of replying to a CSR message so we used this variable to control for the response variability among drivers.

\item\textbf{CSR message length} is the total number of characters in the CSR message. 
\end{itemize}

\noindent Except for the binary and dummy variables, all the numerical control and independent variables were standardized, with a mean of zero and standard deviation of one. Additionally, we took the logarithm of the variables \textit{Days since signup} and \textit{Num of driver messages} before they were standardized since they had a skewed distribution.

\subsection{Analyses and Results}

This analysis seeks to statistically test the effect of the level of politeness and positivity in CSR messages on driver responsiveness and completion of their first trip.  The unit of analysis was a CSR message. Since the same driver might receive multiple CSR messages from his or her CSR, we built random-effects linear regression models which grouped CSR messages at the driver level to deal with non-independence of observations.
The results are shown in Table~\ref{tab:study1} (Model 1).  We omitted the \textit{Sign-up city} variables in the table since there are many of them, and we included them in the models mainly for controlling purposes but not their interpretability.  
 
First, considering the control variables, the driver responsiveness model shows that older drivers and those who were more responsive in previous conversations were more likely to reply to the current message; drivers who had signed up a longer time ago and received longer CSR messages tended not to respond.  On the other hand, except for driver age, all the control variables had a positive effect on the completion of the drivers' first trip.  

Next, we examined the independent variables.  CSR messages with a higher level of politeness were associated with drivers who were more likely to respond and to go through the on-boarding process and complete their first trip.  However, although positive sentiment score was positively correlated with drivers' completing their first trip, it negatively predicts driver responsiveness. The negative association of positivity with driver responsiveness is counter-intuitive.  We found the same negative association when replacing the VADER positive sentiment score with the positive emotion measure from Pennebaker et al.'s Linguistic Inquirey and Word Count program  \cite{pennebaker2001linguistic}, suggesting this result is not a measurement artifact.

\begin{table*}[t]

  \centering
  \begin{tabular}{lrrrr}
    \toprule
    & \multicolumn{2}{c}{Model 1} & \multicolumn{1}{c}{Model 2} & \multicolumn{1}{c}{Model 3} \\
    & \multicolumn{2}{c}{\textbf{(all messages)}} & \multicolumn{1}{c}{\textbf{(no milestone messages)}} & \multicolumn{1}{c}{\textbf{(only question messages)}}\\
    Variable & \multicolumn{1}{c}{Driver Response} & \multicolumn{1}{c}{Driver First Trip} & \multicolumn{1}{c}{Driver Response} & \multicolumn{1}{c}{Driver Response}\\
    \toprule
    Signup city & (omitted) & (omitted) & (omitted) & (omitted)\\
    Driver age & 0.016 *** & -0.001 \space\space\space\space\space\space\space & 0.016 *** & 0.016 *** \\
    Days since signup & -0.105 *** & 0.104 *** & -0.112 *** & -0.148 ***\\
    Num of driver msg & 0.068 *** & 0.053 *** & 0.075 *** & 0.098 ***\\
    CSR msg length & -0.041 *** & 0.010 *** & -0.041 *** & -0.062 ***\\
    \hline
    Politeness & 0.038 *** & 0.001 **\space\space\space & 0.038 *** & 0.016 *** \\
    Positivity & -0.047 *** & 0.001 **\space\space\space & -0.047 *** & -0.004 ***\\
  \bottomrule
  \multicolumn{5}{l}{Coefficients are reported.  *:p<0.05, **:p<0.01, ***:p<0.001}
  \end{tabular}
  
  \caption{Results of the regression analyses.  Model 1 is based on all the data without any filtering.  Model 2 and Model 3 report the analysis results of driver responsiveness where the milestone reminder messages and messages without any question mark are removed, respectively. }
  \label{tab:study1}
  \vspace{3mm}
\end{table*}

\subsubsection{Investigating the negative relationship between CSR positivity and driver responsiveness}

One of our speculations about why positivity of a CSR's utterance was negatively associated with driver responsiveness is that CSRs sent a congratulatory message every time drivers achieved a milestone.  These messages were template-based and crafted with a highly positive tone and thus had a high positive sentiment score (e.g. \textit{``Hi <Name>, This is <Name> from <Company>. Congrats - your background check is complete!''}). However, because they signalled the completion of a subtaks drivers usually did not reply to this kind of status update messages.

To test this speculation, we conducted two additional regression analyses by 1) removing the congratulatory / milestone reminder messages from the data using a set of keywords, such as ``congrats'' and ``congratulations'' and 2) only keeping messages with a question mark which warrant responses.  The first analysis filtered out about 5\% of messages; the second analysis removed 53\% of messages that do not contain any question mark.  The results for the two analyses are presented in Model 2 and Model 3 in Table~\ref{tab:study1}.

Both analyses still show a negative relationships between CSR positive sentiment and driver responsiveness.  We suggest two possible explanations.  First, drivers might actually not care about whether the messages they received were positive or not, but rather want to focus on the task and go through the onboarding funnel as quickly as possible.  They might find the positive messages from agents redundant and thus did not bother to reply.  The second possible explanation is that although positivity has been shown to have a positive impact on user responsiveness in the prior literature, we operationalized it as positive sentiment and measured it with a sentiment analyzer.  Although positive sentiment, which is usually defined more as an endorsement (e.g. like, great, good), the positive/negative connotation of words in a utterance, or polarity of its sentences (e.g. \textit{``remember to ...''} versus \textit{``don't forget to ...''}), may not be a good proxy for positivity (i.e., cheerful or upbeat).  This discrepancy between positivity and positive sentiment might be the reason we found an unexpected effect. Despite this finding, positivity still had a positive association with first trip completion, which is the desired end goal of the on-boarding process. Therefore we retained positivity when building a generative model of social language for an automated agent, as described in Study 2 below.

In sum, the significant impact of politeness and positive sentiment on driver responsiveness and first trip completion is consistent with the thesis that social language in task-oriented conversations can help achieve the task. These findings  motivate us to propose a novel conversational agent framework which can automatically generate agent responses with the desired amount of social language given driver messages as input.

\section{Study 2: A Conversational Agent that Generates Responses with Social Language}

This section presents a conversational agent which can generate responses with more or less social language.  In particular, the ability to adjust the use of politeness and positivity in a conversational agent is essential in the customer service domain, since as shown in the previous section both types of social language have positive impact on task completion and driver's first trip, which is the outcome metric we care the most for the driver on-boarding task.

\subsection{Agent Response Generation}

\begin{figure*}[t]
\centering
\begin{subfigure}[t]{.45\textwidth}
  \centering
  \includegraphics[trim = 40mm 54mm 30mm 0mm, width=1.0\columnwidth]{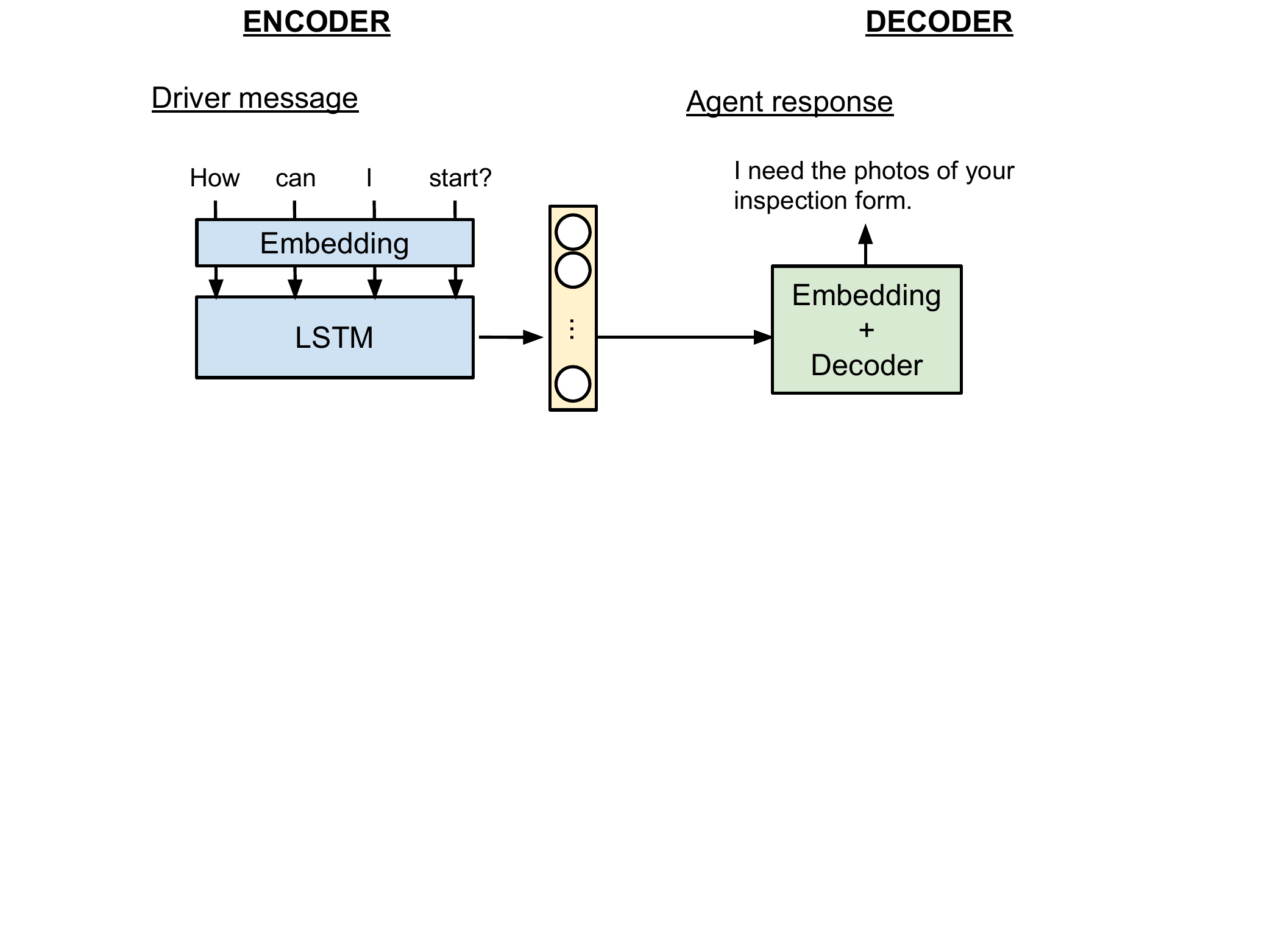}
  \caption{}
  \label{fig:seq2seq}
\end{subfigure}%
\rulesep
\begin{subfigure}[t]{.45\textwidth}
  \centering
  \includegraphics[trim = 25mm 50mm 40mm 2mm, width=1.0\columnwidth]{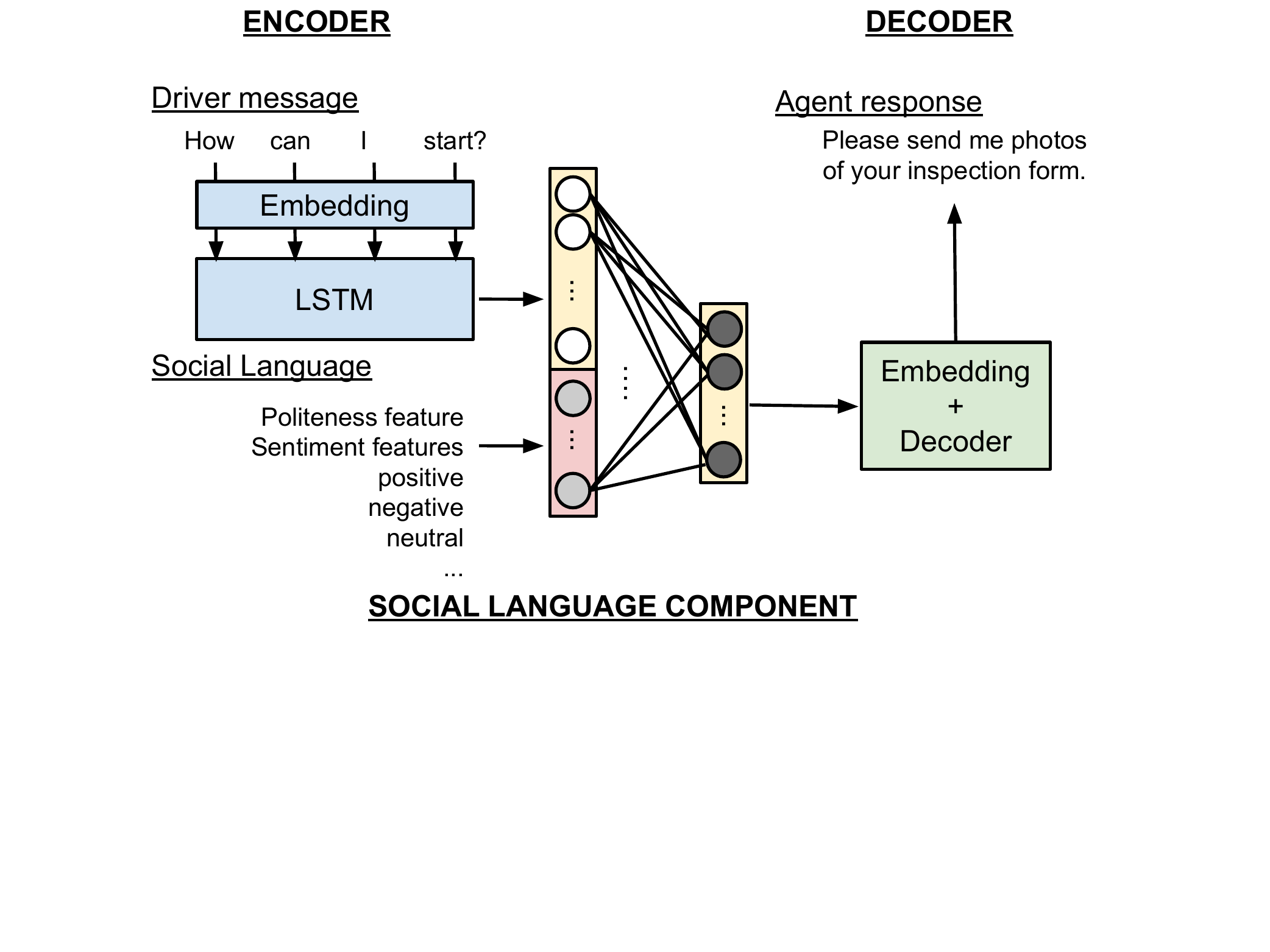}
  \caption{}
  \label{fig:seq2seq_social}
\end{subfigure}

\caption{Model architectures of the baseline and the proposed model: (a) The baseline lexical model implements a typical \textit{seq2seq} architecture.  It has an embedding layer to convert one-hot word representations to dense representations, and utilizes LSTM cells to capture dependencies among words. (b) The proposed model has a social language understanding component in-between the encoder and decoder to integrate the politeness and positivity with the lexical information.}
\label{fig:model}
\end{figure*}

In recent years, deep neural network models have dominated AI research because of their effectiveness.  There is a line of research on developing deep network language generation models to control for a specific linguistic style, including sentiment \cite{shen2017style, hu2017toward, P18-1080}, personality \cite{W18-5019}, and politeness \cite{sennrich2016controlling}.  Language style transfer is the idea of changing the style of a text while preserving its underlying meaning.  Style transfer can be especially useful and relevant to dialogue generation in conversational agents. For example, Oraby et al. \cite{W18-5019} combined the idea of style transfer with a task-oriented dialogue model and have shown that doing so can alter the personality expressed in an output utterance by varying the personality parameters in the input vector.  However, their model was trained and verified on a small synthetic dataset, so its generalizability and practicality is not clear. 

Among all types of deep learning architectures, a sequence-to-sequence learning approach (\textit{seq2seq}) has been most widely and successfully adopted for natural language generation problems, such as machine translation \cite{NIPS2014_5346}, question answering \cite{Yin:2016:NGQ:3060832.3061037}, text summarization \cite{chopra2016abstractive}, and conversational models \cite{DBLP:journals/corr/VinyalsL15, P15-1152, serban2016building}.  A typical \textit{seq2seq} model is designed to transform one sequence to another. To do so, it has two sub-modules: an \textbf{encoder} and a \textbf{decoder}.  The encoder takes a sequence as input and internalizes it as a vector representation, which is then passed to the decoder to generate a corresponding output sequence.  When applied to end-to-end conversational modeling, it generates the next utterance given the previous utterance.  In our case, the input sequence is a driver message, and the output sequence is a CSR response.

To incorporate social language into a \textit{seq2seq} model, we built upon an architecture inspired by Huber et al. \cite{Huber:2018:EDG:3173574.3173851}.  They proposed a conversational agent to generate emotionally appropriate responses by extracting features from images attached in conversations.  In order to integrate visual information extracted from images into a dialogue model, they modified a \textit{seq2seq} structure that uses visual information together with lexical input for conversational language generation.  We modified their architecture by replacing the image understanding layer with a social language understanding component.  The politeness and positivity features are extracted from CSR responses using the pre-trained classifiers described in Study 1.  We evaluated this model against the baseline, a typical \textit{seq2seq} model without the social language component.  Figure~\ref{fig:model} presents the architectures of the baseline and the proposed model, and we describe the models' details below:

\subsubsection{Lexical Model} Our baseline is a classic \textit{seq2seq} model \cite{NIPS2014_5346}, which transforms a driver message to a CSR response.  We added an embedding layer for both the encoder and decoder to convert sparse one-hot word representation (a binary string where only the index of the active word is 1 and all others are 0) to dense vector representation \cite{Mikolov:2013:DRW:2999792.2999959}.  The main advantage of embedding is that it maps words into a latent semantic space so that words with similar meanings and contexts would be closer to each other in that space (e.g. \textit{picture} and \textit{photo}).  We built an encoder and decoder recurrent neural network (RNN) with long short-term memory units (LSTM) so that the model can capture word dependencies \cite{hochreiter1997long}.  The embedding dimension is 300, and the dimensionality of the internal state is set to 512.

\subsubsection{Lexical + Social Model} To introduce social language in agent responses, we introduced a social language understanding component in-between the encoder and decoder as shown in Figure~\ref{fig:model}(b).  During the model training phase, we applied the pre-trained politeness and sentiment classifiers introduced in Study 1 to extract social language features from CSR responses.  We concatenated the social feature vector with the lexical feature vector output by the encoder and passed it to a fully-connected feed-forward neural network.  The output values of the fully-connected layer then become the initial state of the decoder.  We did not employ attention mechanisms or more complex models because our goal is to directly evaluate the impact of the social language layer on the output.

\begin{table*}
  \centering
  \begin{tabular}{l l l}
    \toprule
    Model &\textit{BLEU} score&\textit{word2vec}-based similarity\\
    \toprule
    Lexical Model  & 3.17 & 0.689 \\
    Lexical + Social Model & 9.89 (212\%) & 0.750 (9.38\%) \\
  \bottomrule
  \end{tabular}
  \caption{Results of content preservation evaluation using the \textit{BLEU} score and \textit{word2vec}-based similarity.  The numbers in parentheses refer to the relative increase in the corresponding measure when we compared the Lexical + Social Model to the Lexical Model.}
  \label{tab:study2}
  %\vspace{3mm}
\end{table*}

\begin{table*}
  \centering
  \begin{tabular}{l l l}
    \toprule
    Agent Response &Avg. Politeness Score & Avg. Positivity Score\\
    \toprule
    Unenhanced  & 0.601 & 0.147 \\
    Enhanced & 0.737 & 0.259 \\
     & (22.7\%; p<2.2e-16, t=75.567, df=45317) & (76.6\%; p<2.2e-16, t=74.936, df=45874) \\
  \bottomrule
  \end{tabular}
%\squeezeup
  \caption{Automatic assessment on politeness / positivity between unenhanced and enhanced agent responses.}
  \label{tab:polite_pos_scores}
  \vspace{3mm}
\end{table*}

% train test set

Our dataset consists of 233,571 data points.  Each data point is a pair of a driver message and a CSR response along with the politeness and sentiment scores extracted from the CSR response.  Driver-CSR messages were paired together only if the CSR reply was sent within an hour after the driver's inquiry.  The data was split into train, validation, and test sets with a ratio of 80\%:10\%:10\% where data points that came from the same driver would only be assigned to one set.  We trained both models on the training set with early stopping based on the validation loss.  We to evaluated the agent responses generated by our models using both automatic methods and human judgments.  All the evaluation results reported below were based on the hold-out test set.

\subsection{Evaluation of Content Preservation} 

We conducted automatic evaluations to examine the qualities of the generated text using the \textit{BLEU} score \cite{Papineni:2002:BMA:1073083.1073135} and \textit{word2vec} similarity measure \cite{mikolov2013efficient}.  Both measures consider the text similarity between the actual CSR responses and the model-generated responses.  The difference between them is that the \textit{BLEU} score is a metric based on n-gram overlap while the \textit{word2vec} similarity measures high-level semantic similarity.  Our goal is to quantitatively inspect whether and how much the model-generated responses preserve the content in the ground-truth responses.  The idea is that although we introduced a social language component in the model, we expected that the lexical feature vector output by the encoder should still capture the content, and thus the model should perform at least as well as the baseline model.

\subsubsection{\textit{BLEU} Score}
The \textit{BLEU} score (bilingual evaluation understudy) \cite{Papineni:2002:BMA:1073083.1073135} is a metric originally developed to evaluate the qualities of machine translation models.  Recently, it has also been used to evaluate dialogue generation tasks \cite{N15-1020, N16-1014}.  \textit{BLEU} score, ranging from 0 to 100, is a precision metric that measures how similar a generated response is to the actual human response.  It quantifies the amount of n-gram overlaps between the two.  It also penalizes a generated response that is shorter than the actual response so it does not favor short responses.

\subsubsection{\textit{word2vec} Similarity Measure}
We computed how similar the model outputs are to the ground truth in terms of their \textit{word2vec} \cite{mikolov2013efficient} representations. \textit{word2vec} is one of the state-of-the-art word embedding methods, which convert each word to a vector representation in a latent semantic space such that words used in common contexts are positioned close together in that space.  Specifically, for each utterance, we mapped its words to their word embedding vectors using a \textit{word2vec} model pre-trained on Google News.  Then we averaged the word embedding vectors across the utterance to derive a vector representation for that utterance. We did that for both the model-generated response and its corresponding ground truth (i.e., the CSR's actual response), and computed the cosine similarity between their vector representations as their \textit{word2vec} similarity measure.

We computed the \textit{BLEU} score and \textit{word2vec}-based cosine similarity on a test set of 22,947 pairs of ground truth and model responses.  The results are summarized in Table~\ref{tab:study2}.  We found that adding the social language information significantly improves both measures: 212\% relative increase in \textit{BLEU} and 9.38\% relative increase in \textit{word2vec} similarity (pairwise t-test \textit{p}<.001).  This finding suggests that the social model was better able to preserve the content in the ground-truth responses than the baseline \textit{seq2seq} model even though the social model added a social language understanding component between the encoder and decoder.

\subsection{Social Language Evaluation}

After confirming that the social model can maintain content, the next step is to investigate whether we can adjust the level of politeness or sentiment in the model-generated agent responses by changing the value of the politeness or sentiment feature in the \textit{Lexical+Social} model.  We conducted an automatic analysis on the model outputs using the politeness and sentiment classifiers and also utilized crowdsourcing to collect human judgments on the model-generated dialogue responses.

\subsubsection{Evaluation Setup}

For each driver message in the test set, we generated two agent responses using the \textit{Lexical+Social} model: \textbf{politeness-unenhanced} and \textbf{politeness-enhanced} responses.  The politeness-unenhanced one was generated based on the original level of politeness extracted from the ground truth CSR response; the politeness-enhanced one was produced with the politeness feature value increased by one standard deviation, where the standard deviation of the politeness feature was calculated from the test set.  Using the same approach, we also generated two agent responses for positivity (\textbf{positivity-unenhanced} versus \textbf{positivity-enhanced} responses).

%add some model output example
% and computed the agreement among them.
\begin{table*}[t]

  \centering
  \begin{tabular}{p{13.5cm}}
    \toprule
    \textbf{Driver message}: i need to do the inspection just looking for a place close to my house\\
    \textbf{Agent response 1}: visit any one of these locations for a free inspection - <url> \\
    \textbf{Agent response 2}: ok , i 'll send you a link to the nearest free inspection location .  \\
    \toprule 
    
    Please answer the following three questions:\\

\hspace{10pt}     Q1: Is the agent response 1 reasonable and appropriate to answer the driver message? (Yes/No) \\
\hspace{10pt}     Q2: Is the agent response 2 reasonable and appropriate to answer the driver message? (Yes/No) \\
\hspace{10pt}     Q3: Which response is more polite? (1/2/cannot tell) \\

  \bottomrule
  \end{tabular}
  \caption{The crowdsourcing task for comparing the politeness-unenhanced and politeness-enhanced agent responses.}
  \label{tab:crowdsource_task}
  \vspace{3mm}
\end{table*}

\begin{table*}[!htb]
    \centering
    \begin{subtable}{0.8\linewidth}
      \centering
        
        \begin{tabular}{ll}
            \toprule
            Question & \% of Majority Vote\\
            \toprule
            Q1. Appropriateness (Unenhanced) & 34.0\% \\
            Q2. Appropriateness (Enhanced) & 35.0\% \\
            Q3. More polite (Unenhanced vs Enhanced) & 13.0\% vs 44.0\% (p=1.4e-11, chi2=45.651) \\
            \bottomrule

        \end{tabular}
        \caption{Politeness}
        \vspace{3mm}
    \end{subtable}
    \begin{subtable}{0.8\linewidth}
      \centering
        
        \begin{tabular}{ll}
            
            \toprule
            Question & \% of Majority Vote\\
            \toprule
            Q1. Appropriateness (Unenhanced) & 58.0\% \\
            Q2. Appropriateness (Enhanced) & 62.5\% \\
            Q3. More positive (Unenhanced vs Enhanced) & 35.0\% vs 28.0\% (p=0.162, chi2=1.958) \\
            \bottomrule
            
        \end{tabular}
        \caption{Positivity}
    \end{subtable} 
  \caption{The result of the crowdsourcing task.}
  \label{tab:human_consensus}
  \vspace{3mm}
\end{table*}

\subsubsection{Automatic Analysis}

We applied the politeness and sentiment classifiers to automatically rate both unenhanced and enhanced responses on the entire test set.  Upon doing so, we observed that the responses with the enhanced politeness or positivity input features have significantly higher politeness or positivity scores (Table \ref{tab:polite_pos_scores}).  The results together with the content preservation evaluation provide evidence that the proposed model can generate responses that address drivers' task-oriented concerns with varying degrees of politeness or positivity.

\subsubsection{Human Judgment}

In addition to automatic analysis, we further conducted a pilot study to evaluate the unenhanced and enhanced responses through human judgments.  There were two crowdsourcing tasks, one for politeness, the other for positivity.  Since the two tasks are similar, we will only explain the politeness task here.

In the politeness crowdsourcing task (Table ~\ref{tab:crowdsource_task}), crowdworkers were presented with a driver message and two agent responses generated by the model (the \textbf{politeness-unenhanced} and \textbf{politeness-enhanced} responses) in random order.  Crowdworkers were then asked to answer three questions.  The first two questions were used to assess whether the two generated responses appropriately addressed the driver's message (e.g. answered a question).  We included these two questions to check whether our model can produce a semantically appropriate and reasonable agent response given the driver's prior utterance as context.  The third question asked them to compare the two responses and choose which was more polite.

Although human's perception of the politeness and positivity of an utterance can be influenced by many contextual factors, such as country, culture, dialect, intonation, and personal relationship, the crowdsourcing tasks were designed to measure positivity and politeness at utterance level for simplicity, i.e. from the text of a response without further context apart from the driver's previous message.  Danescu-Niculescu-Mizil et al. \cite{DBLP:conf/acl/Danescu-Niculescu-MizilSJLP13} demonstrated that politeness can be evaluated at the utterance level with a high inter-annotator agreement.  Also, to alleviate the impact of some cultural differences on positivity and politeness judgments, the drivers in our data are all located in US, and the agents and crowdworkers are native English speakers.

We randomly selected 200 data points from the test set and had three crowdworkers make judgments for each politeness task.  There was a similar crowdsourcing task for positivity which was evaluated by  three different crowdworkers.  We measured the inter-rater agreement (\texttt{kappa}) among the three crowdworkers for each question and both tasks \cite{cohen1960coefficient, light1971measures, conger1980integration}.  The result shows that there is a fair to moderate agreement (\textit{Politeness Q1}=.5, \textit{Q2}=.4, \textit{Q3}=.3; \textit{Positivity Q1}=.3, \textit{Q2}=.4, \textit{Q3}=.2).

We evaluated the human judgments by taking a majority vote of the three crowdworkers for each of the three questions. Results are presented in Table \ref{tab:human_consensus}.  For the appropriateness questions (Q1 and Q2), we found no significant difference between the unenhanced and enhanced responses in both the politeness and positivity tasks.  This finding suggests that the content of the generated agent responses, which was controlled by the lexical part of the model, were appropriate to answer drivers' questions and were not affected by the social language component in the \textit{Lexical+Social} model.

Moreover, the analysis of Q3 results indicates that the politeness-enhanced responses were judged as significantly more polite than the unenhanced ones.  In particular, 44\% of the enhanced responses were considered to be more polite than the unenhanced ones, and only 13\% of the unenhanced ones were rated more polite than the enhanced ones.  However, positivity enhancement did not cause crowdworkers to judge the positivity-enhanced responses to be more positive than the unenhanced ones.  This unexpected finding might be due to the small sample or the low agreement for Q3 among crowdworkers in the positivity task (Kappa=.2). The reason that crowdworkers had a low agreement on Q3 might be that the definition of positivity was not clear in the task guidelines.  Another plausible explanation is again that positive sentiment might not a good proxy for postivity. The model was trained with the information provided by the sentiment analyzer while the crowdworkers were asked to compare the output in terms of general positivity.

\section{Conclusion}

\subsection{Discussion}
In this Study 1  we investigated whether and how social language is related to user engagement in task-oriented conversations. We used existing machine learning models to measure politeness and positivity in our analyses.  The results show that the politeness level in CSR messages was positively correlated with driver's responsiveness and completion of their first trip.  We also found that positivity positively predicts driver's first trip, but it has a negative relationship to driver responsiveness even after removing congratulatory milestone messages or messages that do not have any question mark, which usually have positive sentiment and/or do not require responses from drivers. 

To integrate the findings from the statistical analyses into a dialogue model, Study 2 proposed and evaluated a task-oriented conversational agent model that can generate agent responses with the desired level of politeness / positivity by inserting a social language understanding component into a typical \textit{seq2seq} model.  The automatic evaluations demonstrate that the proposed model can manipulate the politeness or positivity level of agent responses while preserving content.  However, the crowdsourcing evaluation shows that the model can enhance politeness but not positivity in the generated agent responses.  A common explanation for the negative association of positivity with driver responsiveness in Study 1 and the lack of an effect of positivity enhancement on generated agent responses in Study 2 might be a discrepancy between the concept of language positivity and its  operationalization as positive sentiment. That is, the results may reflect a mismatch between what we thought we were measuring and manipulating and what we  actually measured and manipulated.

The contributions in this research were  1) using off-the-shelf classifiers as a way to detect social language behavior for quantitative analyses and 2) incorporating them in a task-oriented conversational agent framework.

\subsection{Implications}

This work has several implications.  First, although we only focus on politeness and positivity, we believe that our proposed modeling framework should be extended to incorporate other kinds of types of social language into task-oriented assistants.  The model can also be used to provide drivers with better experience during their on-boarding process.  The customer support services can be improved by utilizing the model to provide suggested replies to CSRs so that they can (1) respond quicker and (2) adhere to the best practices (e.g. using more polite and positive language) while still achieving the goal that the drivers and the ride-sharing providers share, i.e., getting drivers on the road.

\subsection{Limitations and Future Directions}

% Need MORE crowdworkers!
The moderate inter-rater agreement of the crowdsourcing task and the unexpected human evaluation results for positivity suggest that crowdworkers might have been confused  about the task.  Follow-up research should investigate the labelling process and refine the task.  Specifically, one suggestion we received from the crowdworkers is to provide more context in terms of conversation history to them so they can make better judgments.  We would also like to expand our crowdsourcing effort to engage more workers to review and evaluate more model-generated responses.

%positivity classifier
Since we suspect there might be some difference between positivity and positive sentiment, one future research direction is to build a positivity machine learning model so we can measure the concept directly.

% A/B test
Finally, although we found that there is a correlation between social language and user engagement, the results are based on the analysis of human-human conversations.  We need to validate whether this finding also applies to human-bot interactions, i.e., whether a polite conversational agent would also improve user engagement.  Future work should conduct A/B tests to examine the effectiveness of a polite and positive conversational agent.

% \section{Acknowledgments}
% This is acknowledgments.

% Balancing columns in a ref list is a bit of a pain because you
% either use a hack like flushend or balance, or manually insert
% a column break.  http://www.tex.ac.uk/cgi-bin/texfaq2html?label=balance
% multicols doesn't work because we're already in two-column mode,
% and flushend isn't awesome, so I choose balance.  See this
% for more info: http://cs.brown.edu/system/software/latex/doc/balance.pdf
%
% Note that in a perfect world balance wants to be in the first
% column of the last page.
%
% If balance doesn't work for you, you can remove that and
% hard-code a column break into the bbl file right before you
% submit:
%
% http://stackoverflow.com/questions/2149854/how-to-manually-equalize-columns-
% in-an-ieee-paper-if-using-bibtex
%
% Or, just remove \balance and give up on balancing the last page.
%
\balance{}

% BALANCE COLUMNS
\balance{}

% REFERENCES FORMAT
% References must be the same font size as other body text.
\bibliographystyle{SIGCHI-Reference-Format}
\bibliography{sample}

\end{document}